\pdfoutput=1
\documentclass{article}
\usepackage{acl}
\usepackage{times}
\usepackage{latexsym}
\usepackage[T1]{fontenc}
\usepackage[utf8]{inputenc}
\usepackage{microtype}
\usepackage{xcolor}
\usepackage{tcolorbox}
\usepackage{inconsolata}
\usepackage{graphicx}
\usepackage{amsmath}
\usepackage{hyperref}
\usepackage{booktabs}
\usepackage{multirow}
\usepackage{amssymb}
\usepackage{enumitem}
\setlist[itemize,enumerate]{leftmargin=*}
\title{Object-Level Verbalized Confidence Calibration in Vision-Language Models via Semantic Perturbation}

\author{
    \textbf{Yunpu Zhao\textsuperscript{1}},
    \textbf{Rui Zhang\textsuperscript{2}},
    \textbf{Junbin Xiao\textsuperscript{3}},
    \textbf{Ruibo Hou\textsuperscript{5}},
    \\
    \textbf{Jiaming Guo\textsuperscript{2}},
    \textbf{Zihao Zhang\textsuperscript{2}},
    \textbf{Yifan Hao\textsuperscript{2}},
    \textbf{Yunji Chen\textsuperscript{2,3}},
    \\
    \textsuperscript{1}University of Science and Technology of China
    \\
    \textsuperscript{2}SKL of Processors, Institute of Computing Technology, CAS
    \textsuperscript{3}National University of Singapore
    \\
    \textsuperscript{4}University of Chinese Academy of Sciences
    \textsuperscript{5}University of Illinois at Urbana-Champaign
    \\
}

\begin{document}

\maketitle

\begin{abstract}
    Vision-language models (VLMs) excel in various multimodal tasks but frequently suffer from poor calibration, resulting in misalignment between their verbalized confidence and response correctness. This miscalibration undermines user trust, especially when models confidently provide incorrect or fabricated information.
    In this work, we propose a novel Confidence Calibration through Semantic Perturbation (CSP) framework to improve the calibration of verbalized confidence for VLMs in response to object-centric queries.
    We first introduce a perturbed dataset where Gaussian noise is applied to the key object regions to simulate visual uncertainty at different confidence levels, establishing an explicit mapping between visual ambiguity and confidence levels.
    We further enhance calibration through a two-stage training process combining supervised fine-tuning on the perturbed dataset with subsequent preference optimization. 
    Extensive experiments on popular benchmarks demonstrate that our method significantly improves the alignment between verbalized confidence and response correctness while maintaining or enhancing overall task performance. 
    These results highlight the potential of semantic perturbation as a practical tool for improving the reliability and interpretability of VLMs.
\end{abstract}

\section{Introduction}
\begin{figure}
    \centering
    \includegraphics[width=1\linewidth]{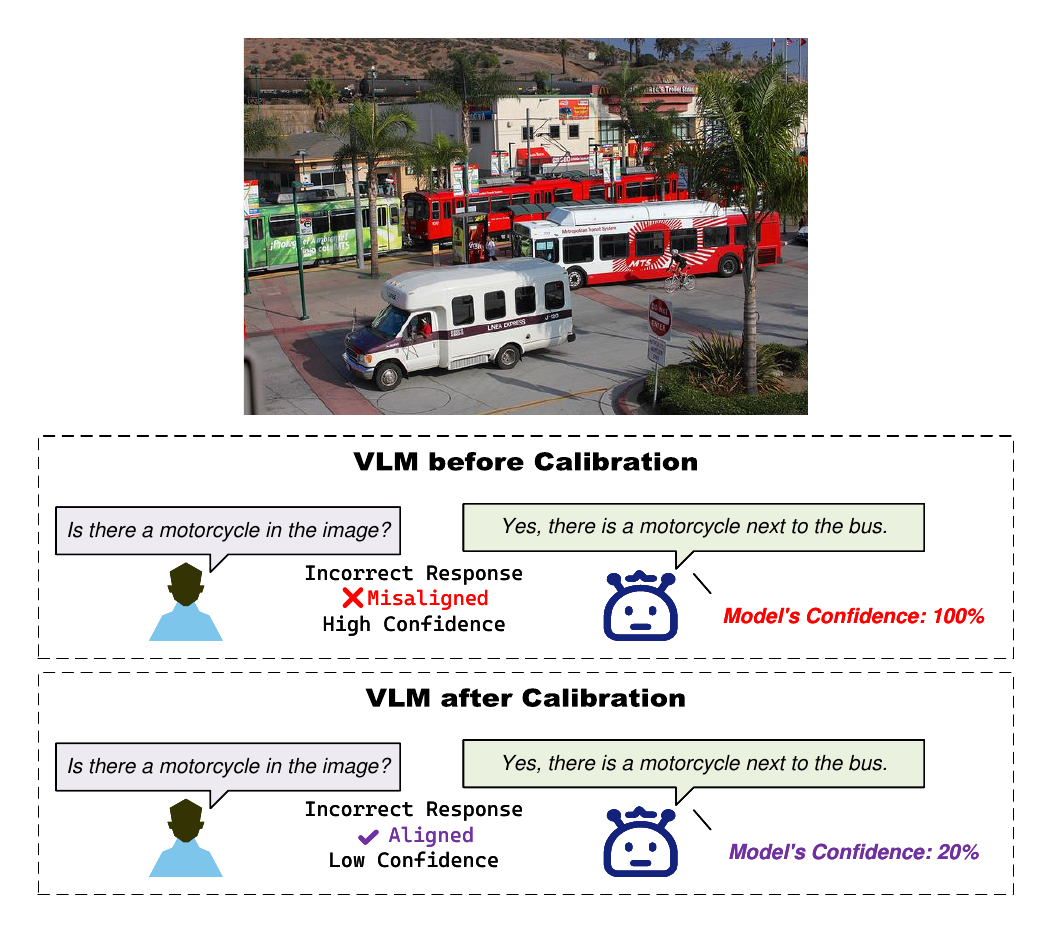}
    \caption{(Top): Most current VLMs tend to generate high verbalized confidence on incorrect response. (Bottom): After calibration, model's verbalized confidence will be aligned with response correctness.}
    \label{example1}
\end{figure}

Modern vision-language models (VLMs) have demonstrated remarkable success on tasks ranging from image captioning to visual question answering \cite{Gpt-4,Sparks_of_AGI}. \cite{Gpt-4,Sparks_of_AGI}. 
In safety‑critical or user‑facing settings, however, accuracy alone is not enough; users also need to know how sure the model is about each object it claims to see. Verbalized confidence (e.g., “I’m 80\% sure there is a cat”) lets downstream systems weight or reject predictions according to their stated reliability.
Unfortunately, today’s VLMs frequently display object‑level overconfidence: they insist a non‑existent object is present with near‑100\% certainty, or hedge only weakly when the visual evidence is ambiguous (Fig. \ref{example1}, top). 
Such misaligned confidence is especially hazardous in object‑centric applications like product‑search, autonomous driving, or medical triage, where a single hallucinated object can trigger costly decisions. Calibrating verbalized confidence—so that high confidence accompanies truly correct object detections while low confidence flags likely hallucinations (Fig.\ref{example1}, bottom)—is therefore essential for trustworthy deployment.

Text‑only calibration methods \cite{logit_based_1,verbalization_based_1} transfer poorly to this multimodal, object‑focused setting, because VLMs face two extra hurdles. \textbf{(i) Local visual uncertainty:} occlusion, blur, and lighting often hide or distort the very region that disambiguates an object \cite{VLM-Uncertainties}. 
\textbf{(ii) Multimodal bias:} large language priors can dominate scarce visual evidence, causing the model to “talk itself into” seeing an object that is not there \cite{sycophancy}. 
Consequently, due to the combined effects of introduced visual uncertainty and multimodal imbalance, these factors keep object‑level verbalized‐confidence calibration an unsolved issue.

We tackle this gap with Calibration through Semantic Perturbation (CSP), a novel framework explicitly designed to calibrate verbalized confidence for object-level queries for VLMs. 
We introduce a perturbed dataset that modifies key visual elements based on different confidence levels, allowing the model to learn explicit mappings between visual uncertainties and verbalized confidence.
To construct the perturbed dataset, we first extract key object regions referenced in a multimodal query using GroundingDINO \cite{GroundingDINO} and Segment Anything (SAM) \cite{SAM}. 
We then apply varying levels of Gaussian noise to these regions that mimic different degrees of occlusion or distortion. Each perturbed image is associated with a ground-truth confidence label so that the model can learn a direct mapping from how visually reliable the object pixels are → how much confidence it should verbalize.
After that, a two‑stage pipeline—supervised fine‑tuning on these perturbed samples, followed by preference optimization—teaches VLMs to down‑weight hallucinated objects and up‑weight true detections.
In this way, our approach accounts for visual uncertainties in semantic extraction and encourages more visually-grounded confidence judgments, leading to sharply improved, object‑aligned verbalized confidence.

We conduct extensive experiments on widely-used VLM benchmarks focusing object-level queries across multiple state-of-the-art VLMs. 
Comparing their performance before and after calibration using our CSP framework, our results demonstrate substantial improvements in verbalized confidence calibration across a diverse set of evaluation metrics. 
These improvements show that models trained with CSP correlate their expressed confidence more faithfully with actual correctness, reducing the likelihood of misleadingly high-confidence errors. Moreover, these calibration improvements do not come at the cost of task performance. CSP preserves or enhances the VLMs’ task accuracy, confirming that our method improves trustworthiness and interpretability without sacrificing predictive capability.
In addition, we conduct comprehensive ablation studies to isolate the contribution of each component in our framework. We also observe that CSP improves the consistency between internal confidence scores and verbalized confidence, suggesting that better visual uncertainty modeling can simultaneously benefit both internal and external calibration signals.

In summary, our primary contributions are: 
\begin{enumerate}[topsep=0pt,itemsep=2pt,leftmargin=1em] 
\item \textbf{Novel Calibration Framework.} 
We introduce CSP, a new framework for training VLMs for better object-level calibration with verbalized confidence and response correctness. 
\item \textbf{Semantic Perturbation Data Construction.} 
We present a systematic approach to local image perturbation that simulates diverse levels of visual ambiguity, enabling more fine-grained calibration during training. 
\item \textbf{Extensive Validation.} Empirical results on multiple benchmarks and model architectures show that CSP significantly reduces calibration error and improves trustworthiness, without sacrificing task accuracy. \end{enumerate}

\section{Related Work}
\paragraph{LLM Confidence Calibration.} Confidence calibration has emerged as a critical challenge in LLMs to ensure reliable and trustworthy outputs. Early methods predominantly focused on calibrating internal confidence derived from model logits.
These logit-based approaches often employ statistical techniques like temperature scaling or model-based re-calibration \cite{rw1,rw2} to adjust probability distributions and mitigate overconfidence.
More recent research extends beyond simple scaling by exploiting semantic features and contextual cues to better align token-level probabilities with actual prediction correctness \cite{rw3}.
In particular, linguistic uncertainty modeling and sequence likelihood calibration have shown promise for managing generation tasks susceptible to compounding errors \cite{rw4}.
Although effective in text-only settings, these calibration techniques largely overlook the additional complexity introduced by visual or other multimodal inputs.

\paragraph{Verbalized Confidence.} While most traditional calibration work centers on the gap between predicted probabilities and true correctness, a growing line of research emphasizes the verbalized confidence explicitly stated by the model \cite{verbalization_based_1,confidence-under-hood}. Such verbalized confidence can help users better gauge the reliability of model outputs.
Recent studies have explored prompting strategies and self-reflection pipelines that encourage LLMs to articulate their confidence levels, thereby improving transparency \cite{rw7,rw8}.

\paragraph{Towards Multimodal Calibration.} Despite the progress in unimodal confidence calibration and verbalized confidence expression, few frameworks systematically address these issues in VLMs.
Recent work highlights the need for calibration techniques tailored to multimodal data, which present unique challenges such as object occlusion and semantic ambiguity \cite{Calibration_LLM_Survey,rw5,rw6}.
In contrast to existing methods that primarily focus on textual uncertainty or straightforward logit manipulation, our approach introduces semantic mask perturbation to simulate varying degrees of visual uncertainty, laying the groundwork for more trustworthy multimodal systems.

\section{Methodology}
In this section, we formally define the problem of object-level verbalized confidence calibration, and present our proposed CSP framework. 
As illustrated in Figure~\ref{example2}, CSP consists of two stages: object-centric dataset construction and training. 
In the dataset construction stage, we generate visually perturbed images by systematically applying varying degrees of Gaussian noise to the object regions explicitly referenced in the input query. Each perturbed sample is then associated with a target confidence label, reflecting the intended level of visual uncertainty over the object.
In the training stage, we use the constructed dataset to fine-tune the model via supervised learning, followed by preference optimization to further enhance verbalized confidence alignment. 
Together, these components form a cohesive framework tailored for object-level confidence calibration, enabling VLMs to produce more trustworthy, interpretable, and uncertainty-aware predictions in response to object-centric queries.
\subsection{Problem Definition of Verbalized Confidence Calibration}
The goal of verbalized confidence calibration is to enable a VLM to assign a verbalized confidence score to a candidate answer that accurately reflects the probability of correctness. 

In other words, the correct answer should be assigned with the highest confidence score, while the other incorrect answers should be assigned with lower scores. 

Specifically, given a visual input $v_0$ and a textual query $q$, considering there exists a set of candidate answers \(\ \{a_1, a_2, \dots, a_k\}\), the VLM will assign a verbalized confidence score $c(a_i)$ to each candidate response $a_i$.
For the correct answer $a_*$, verbalized confidence calibration aims to enable the VLM to increase the verbalized confidence score $c(a_*)$ to be the highest, while decrease the verbalized confidence score of the other incorrect answers to be low.
In this way, the objective of generating verbalized confidence score that accurately reflects the correctness of the VLM's answer can be achieved. 

To evaluate the effectiveness of verbalized confidence calibration, we assess whether the answer $\hat{a}$ with highest verbalized confidence score matches the correct answer $a_*$ with a metric $M$, namely calculate the similarity $M(a_*, \hat{a})$.
Here, $M$ can be any commonly used similar metrics, such as Accuracy, F1 Score, Area Under the Curve (AUC), Brier Score, Expected Calibration Error (ECE).
This evaluation jointly reflects both the model's ability to identify correct object-level answers and the fidelity of its expressed confidence.

\subsection{Dataset Construction with semantic perturbation}

\begin{figure*}[tbp]
    \centering
    \includegraphics[width=1\linewidth]{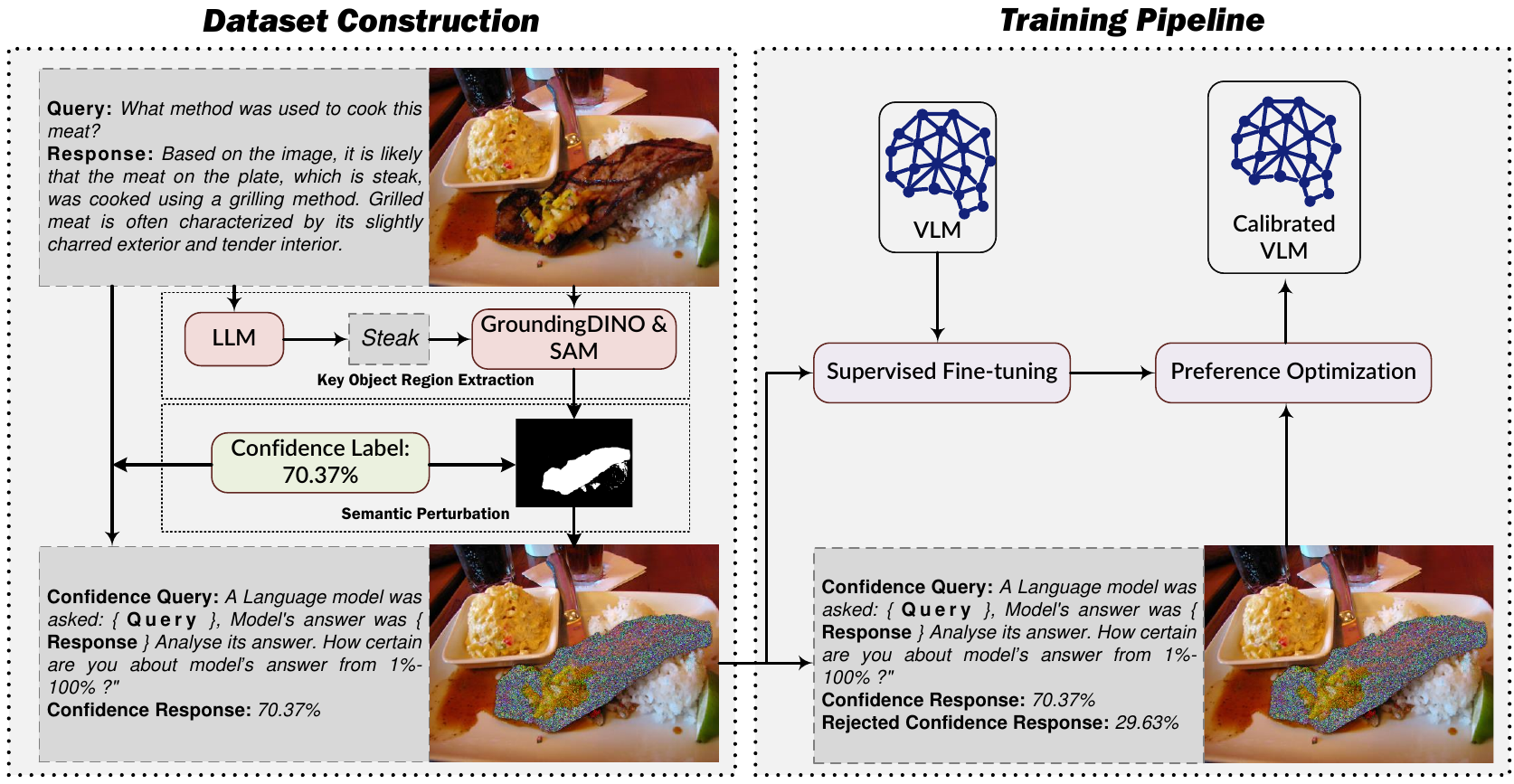}
    \caption{The image illustrates the dataset construction and training pipeline for improving confidence calibration in VLM. It highlights the two-stage process: \textbf{Dataset Construction}: Extracting key object regions using GroundingDINO and SAM, applying semantic perturbations, and assigning confidence labels based on noise levels. \textbf{Training Pipeline}: Fine-tuning the VLM with supervised learning, followed by preference optimization, to improve probability-confidence alignment and response calibration.}
    \label{example2}
\end{figure*}

To address the challenges of verbalized confidence calibration in VLMs, we construct a specialized dataset incorporating semantic perturbations. This design is motivated by two key challenges that hinder accurate confidence expression in VLMs, especially in response to object-level queries.
First, visual uncertainty arises from factors such as occlusion or poor lighting, which obscure key objects and lead to incomplete semantic extraction. Existing training data often lack explicit examples reflecting such conditions, making it difficult for VLMs to properly adjust their confidence when faced with ambiguous object regions.
Second, VLMs exhibit a strong reliance on textual priors while often overlooking critical visual details, creating a multimodal imbalance. Without explicit guidance, VLMs tend to verbalize confidence based primarily on linguistic patterns, which leads to misleading confidence estimates when the visual evidence for object presence is weak or ambiguous.
To mitigate this issue, we construct a dataset where varying levels of perturbations are applied to object regions, conditioned on target confidence levels.
The perturbations are applied selectively to semantically relevant objects or regions rather than indiscriminately across the entire image, ensuring that verbalized confidence is directly grounded in affected visual features.
Thus, this dataset provides a more structured learning framework for addressing verbalized confidence calibration in VLMs for hallucination-prone object recognition tasks..

As illustrated in Figure~\ref{example2}, given a text query and a corresponding image, we first identify the most relevant object mentions and localize their regions using GroundingDINO and SAM. 
We then inject varying intensities of Gaussian noise into the segmented object region, creating a series of perturbed images that mimic different levels of visual uncertainty according to a confidence label. The confidence label is sampled from 0\% to 100\%. 
Finally, we convert each sample into a new confidence query that consists of the original question-answer pair, and a target confidence response corresponding to the perturbation level. By covering diverse noise levels and confidence targets, the expanded dataset forms the foundation for both supervised fine-tuning and preference optimization, ultimately improving the model’s ability to align its verbalized confidence with true visual uncertainty in object-centric tasks.

\paragraph{Key Object Region Extraction:} 
To construct the specialized dataset, we begin with extract the key object region that are relevant to the input query. 
This process consists of three main steps: key object descriptions extraction, object localization and semantic segmentation. 
Given a textual query $q$ and a response $r$, first we extract key object descriptions $M_{\text{desc}}(q, r)$ by using a small LLM, for example, the word ``steak'' shown in the image. 
Then based on the extracted key object descriptions, we use GroundingDINO to localize the most relevant objects in the image that correspond to terms mentioned in the multimodal query-response pair. Given the image $v_0 \in \mathbb{R}^{H \times W}$, GroundingDINO outputs object bounding boxes that approximate the regions of interest.
Next, we apply SAM to refine these localized object regions into precise semantic masks. Given the bounding boxes provided by GroundingDINO, SAM generates pixel-wise segmentation masks that more accurately delineate the object’s shape and boundaries. The final binary mask $m$ is computed as:
\[
m = \text{SAM}(v_0, \text{GroundingDINO}(v_0, M_{\text{desc}}(q,r))),
\]
where \( m \in \{0,1\}^{H \times W} \) denotes the binary mask of size \( H \times W \) which is same as the visual input $v_0$. By combining object detection with fine-grained segmentation, we ensure that only the most relevant semantic regions are perturbed in the next step, allowing for precise control over visual uncertainty in the dataset.

\paragraph{Confidence Labeling and Semantic Perturbation Mechanism:} 
To simulate visual uncertainty in a controlled way, we apply Gaussian noise only to the object regions referenced in the query, as identified by the mask $m$. Let $c\in[0,100]\%$ be the desired confidence label, where $c=100\%$ indicates no perturbation and $c=0\%$ indicates maximal noise. 
In this context, a higher confidence corresponds to lower uncertainty, so we inject less (or no) noise. Conversely, a lower confidence corresponds to higher uncertainty, so more noise is applied to the key object regions. 
Therefore, by decreasing confidence from $100\%$ to $0\%$, we increase the noise to simulate a progressive increase in visual uncertainty.

Following the forward diffusion process in image generation \cite{diffusion}, we inject Gaussian noise to the key object region by mapping $c$ to a diffusion step $T_c$ by a linear schedule:
\[
T_{\mathit{c}} \;=\; \bigl\lfloor T_{\max} \times \bigl(1 - \frac{\mathit{c}}{100}\bigr)\bigr\rfloor,
\]
where \(T_{\max}\) is a chosen upper bound on the number of diffusion steps. Starting from the original image \(v_0\), we iteratively sample:
\[
v_{t} \;\sim\; \mathcal{N}\!\Bigl(\sqrt{1-\gamma}\,v_{t-1},\;\gamma\,\mathbf{I}\Bigr)
\quad\text{for}\quad
t = 1 \dots T_{\mathit{c}},
\]
with \(\mathcal{N}\)denotes a Gaussian distribution. \(\gamma\) is a predefined parameter that controls the noise intensity introduced in each step. A larger $\gamma$ results in stronger noise injection per step, leading to a more rapid degradation of the image. 
This diffusion-based approach enables a gradual and controlled degradation of visual features, allowing for a continuous mapping between the confidence label and the level of perturbation.
After \(T_{\mathit{c}}\) iterations, we combine the noised image \(v_{T_{\mathit{c}}}\) with the unperturbed background of \(v_0\) using the binary mask \(m\):
\[
v_{\text{perturbed}} \;=\; m \,\odot\, v_{T_{\mathit{c}}} \;+\; \bigl(1 - m\bigr)\,\odot\, v_0,
\]
where \(\odot\) denotes element-wise multiplication. 

This procedure distorts only the object region relevant to the given query-response pair, as illustrated in Figure~\ref{example2} (where the region of the object ``steak'' is partially occluded, while the plate and background remain clear).  
Each confidence label \(\mathit{c}\) thus produces a distinct visually perturbed image, ranging from minimal to severe noise.  
By pairing these images with the appropriate textual instructions (e.g., “How certain are you about the model’s answer from 1\% to 100\%?”), we give the model explicit supervision on how to verbalize confidence. This enables the model to associate different degrees of object visibility with corresponding confidence levels, supporting more grounded and interpretable object-level verbalized calibration.

\paragraph{Dataset Integration:} 
The resulting dataset is constructed through dataset modification and augmentation of the RLAIF dataset \cite{rlaif}, which is a large-scale multimodal AI feedback dataset collected from a diverse sources.
We enhance the original dataset by applying the proposed semantic perturbation technique, generating diverse image-query-response samples. 
Each sample in the dataset consists of a transformed query \( q_c \) and a corresponding confidence-labeled answer \( r_c \). 
The transformed $q_c$ is generated from the original query $q$ and response $r$.
To ensure objective and generalized confidence evaluation, we transform \( q_c \) in a Third-Person Perspective (TPP) format, framing the query as an external assessment rather than a direct model introspection, as shown in Figure \ref{example2}.
The corresponding confidence-labeled answer \( r_c \) is assigned as: $r_c = c$ where \( c \) is the confidence label derived from the semantic perturbation process. 
Finally we construct modified semantic perturbed dataset \( D = \{(v_{\text{perturbed}}, q_c, r_c)\} \). We found that first-person confidence queries often resulted in inflated self-assessments. Framing queries in third-person promotes more objective calibration by discouraging sycophantic behavior \cite{confidence-under-hood}.
By simulating real-world visual uncertainties, this dataset is capable of supporting a robust framework for addressing verbalized confidence calibration challenges in vision-language models.

\subsection{Training}

\paragraph{Supervised Fine-tuning (SFT)}
Using the constructed dataset \( D = \{(v_{\text{perturbed}}, q_c, r_c)\} \), we perform SFT to establish the model's capability to associate visual uncertainty with verbalized confidence. The objective of SFT minimizes the cross-entropy loss:
\[
\mathcal{L}_{\text{SFT}} = - \mathbb{E}_{(v_{\text{perturbed}}, q_c, r_c) \sim D} \big[ \log P_\theta(r_c \mid v_{\text{perturbed}}, q_c) \big],
\]
where \( P_\theta(r_c \mid v_{\text{perturbed}}, q_c) \) is the model's probability of generating the response given the visual and textual inputs. 
This step enables the model to learn the relationship between visual uncertainty (as affected by diffusion noise) and verbalized confidence.

\paragraph{Preference Optimization}
To further refine the model’s verbalized confidence calibration, we adopt SimPO (Simple Preference Optimization)~\cite{SIMPO} on top of the SFT model. Specifically, for each training example in the perturbed dataset \(\{(v_{\text{perturbed}}, q_c, r_c)\}\), we take $r_c$ as the winning response and define the rejected response $r_{\text{rej}}=100\%-c$. This pairwise preference setting encourages the model to produce confidence estimates that more closely reflect the visual uncertainty.
Formally, let $\pi_\theta$ denote the policy model and $(q_c,y_w,y_l)$ be a preference sample in which $y_w\equiv r_c$ and $y_l\equiv r_{\text{rej}}$. SimPO optimizes the following margin-based objective:
\[
\small
\begin{aligned}
&\mathcal{L}_{\text{SimPO}}(\pi_\theta) = - \mathbb{E}_{(x,y_w,y_l) \sim D} \\ 
&\Bigg[ \log \sigma \Bigg( \frac{\beta}{|y_w|} \log \pi_\theta(y_w \mid x) - \frac{\beta}{|y_l|} \log \pi_\theta(y_l \mid x) - \lambda \Bigg) \Bigg],
\end{aligned}
\]
where $\sigma$ is the sigmoid function, $\beta$ is a scaling factor for the reward, $\lambda$ is a target margin ensuring the policy assigns sufficiently higher probability to the winning response compared to the losing response.
In conjunction with the SFT step, this preference-based fine-tuning further ensures that the final model’s verbalized confidence provides a faithful reflection of the true visual uncertainty in the input.

\section{Experiments}
\subsection{Experimental Settings}
\paragraph{Dataset}
We conduct experiments on two popular datasets POPE \cite{POPE} and AMBER \cite{AMBER} to verify the effectiveness of our proposed method. POPE, the Polling-based Object Probing Evaluation, is designed to assess object hallucination in VLMs regarding the presence of objects in images. POPE is divided into three settings: random, popular, and adversarial, indicating different methods of sampling hallucination objects. AMBER is a comprehensive benchmark designed to evaluation multiple different types of hallucination including attribute hallucination and relation hallucination. We choose hallucination benchmarks to validate verbalized confidence calibration because calibration tends to be more challenging in scenarios with severe hallucination, making these benchmarks particularly representative for assessing the effectiveness of our approach.
\paragraph{Models}
We benchmarked the proposed method against several state-of-the-art vision-language models: Qwen-VL-Chat \cite{Qwen-VL}, Qwen2-VL-7B-Instruct \cite{Qwen2-VL}, InternVL2-8B \cite{InternVL}, and Phi-3.5-vision-instruct \cite{phi-3.5}. Qwen-VL is a versatile vision-language model adept at understanding, localization, and text reading tasks. Qwen2-VL, an advanced iteration of Qwen-VL, enhances image comprehension across various resolutions and ratios, and extends capabilities to video understanding and multilingual support. InternVL2 is an open-source multimodal large language model designed to bridge the gap between open-source and proprietary commercial models in multimodal understanding. Phi-3.5 is a vision-language model that provides general-purpose AI capabilities, handling both visual and textual inputs efficiently. For each baseline, we utilized the official pre-trained models and followed the recommended evaluation protocols to ensure a fair comparison. For more detailed information on the experimental configuration, please refer to the appendix.

\subsection{Evaluation Metrics}
We employ five metrics to evaluate our model’s verbalized confidence calibration.
Let \(\hat{y_i}\) be the prediction chosen by the highest confidence \( c(\hat{y_i}) \), \(y_i\) be the ground-truth, and $p_i\in[0,1]$ denote the corresponding verbalized confidence score served as the soft probability to compute confidence-aware metrics.
We define:
\begin{itemize}[topsep=0pt,itemsep=2pt,leftmargin=1em]
    \setlength{\itemsep}{0pt} 
    \setlength{\parskip}{0pt} 
    \setlength{\parsep}{0pt}
    \item \textbf{Accuracy (Acc)}: 
    \(\displaystyle
       \text{Acc} = \frac{1}{N}\sum_{i=1}^{N} \mathbb{I}(\hat{y}_i = y_i).
    \)
    \item \textbf{F1 Score}: 
    The harmonic mean of precision and recall.
    \item \textbf{AUC (Area Under ROC Curve)}: 
    The probability that a randomly chosen correct instance is ranked higher (by \( c(\hat{y_i}) \)) than an incorrect instance.
    \item \textbf{Brier Score (BS)}: 
    \(\displaystyle
      \text{BS} = \frac{1}{N}\sum_{i=1}^{N} \bigl(p_i - y_i\bigr)^2,
    \)
    where \(p_i \in [0,1]\) is the model’s predicted probability (verbalized confidence) for the event \(y_i=1\).
    \item \textbf{Expected Calibration Error (ECE)}: 
    Partition samples into \(K\) bins of equal confidence range; measure the average gap between mean predicted confidence and empirical accuracy in each bin:
    \[
      \text{ECE} = \sum_{k=1}^{K} \frac{|B_k|}{N} \bigl|\text{acc}(B_k) - \text{conf}(B_k)\bigr|.
    \]
    Here, \(\text{acc}(B_k)\) is the average correctness and \(\text{conf}(B_k)\) is the average predicted confidence in bin \(B_k\).
\end{itemize}
Lower BS, ECE and higher Acc, F1, AUC indicate better calibration and overall alignment of confidence with correctness. 
These metrics are complementary: Acc and F1 evaluate how reliably the model's most confident predictions align with correctness, while AUC, BS, and ECE assess distinct facets of calibration—ranking reliability across all confidence thresholds, probability sharpness, and bin-wise confidence-accuracy alignment, respectively. 
Together, they provide a holistic view of verbalized confidence calibration, measuring both the trustworthiness of confidence-guided predictions and the statistical alignment between expressed confidence and empirical correctness.

\subsection{Experimental Results}
To evaluate the effectiveness of our proposed CSP framework, we assess the calibration of object-level verbalized confidence across multiple datasets according to the metrics we introduced above.
\subsubsection{Results of Verbalized Confidence Calibration}
Table~\ref{Calibration} demonstrates significant improvements in calibration across all models and datasets. Our method enhances accuracy and F1 score while reducing ECE, indicating better confidence correctness alignment. Notably, Qwen2-VL, which initially exhibited severe object-level overconfidence, shows substantial improvement after training with CSP. Conversely, InternVL2, a relatively well-calibrated model, still benefits from CSP, demonstrating enhanced robustness under adversarial and ambiguous visual conditions. These results highlight that our approach not only improves the ability of verbalized confidence prediction in weaker models but also refines confidence estimation in stronger ones, making VLMs more reliable and interpretable when answering object-centric queries under visual uncertainty.
\begin{table*}[!t]
\centering
\resizebox{\textwidth}{!}{%
\begin{tabular}{@{}cccccccccccccccc@{}}
\toprule
Model & \multicolumn{9}{c}{POPE} & \multicolumn{6}{c}{AMBER} \\ \midrule
 & \multicolumn{3}{c}{Random} & \multicolumn{3}{c}{Popular} & \multicolumn{3}{c}{Adversarial} & \multicolumn{3}{c}{Attribute} & \multicolumn{3}{c}{Relation} \\
 & \textit{Acc}~(\(\uparrow\)) & \textit{F1}~(\(\uparrow\)) & \textit{ECE}~(\(\downarrow\)) & \textit{Acc}~(\(\uparrow\)) & \textit{F1}~(\(\uparrow\)) & \textit{ECE}~(\(\downarrow\)) & \textit{Acc}~(\(\uparrow\)) & \textit{F1}~(\(\uparrow\)) & \textit{ECE}~(\(\downarrow\)) & \textit{Acc}~(\(\uparrow\)) & \textit{F1}~(\(\uparrow\)) & \textit{ECE}~(\(\downarrow\)) & \textit{Acc}~(\(\uparrow\)) & \textit{F1}~(\(\uparrow\)) & \textit{ECE}~(\(\downarrow\)) \\ \midrule
Qwen-VL & 0.25 & 0.21 & 0.5699 & 0.3 & 0.22 & 0.5249 & 0.27 & 0.21 & 0.5475 & 0.37 & 0.47 & 0.4732 & 0.1 & 0.07 & 0.4421 \\
\textit{using CSP} & \textbf{0.67} & \textit{\textbf{0.68}} & \textbf{0.4225} & \textbf{0.67} & \textbf{0.68} & \textbf{0.4289} & \textbf{0.61} & \textbf{0.64} & \textbf{0.4407} & \textbf{0.69} & \textbf{0.7} & \textbf{0.4158} & \textbf{0.6} & \textbf{0.7} & \textbf{0.3674} \\
Qwen2-VL & 0.11 & 0.01 & 0.412 & 0.04 & 0.01 & 0.477 & 0.04 & 0.01 & 0.4767 & 0.13 & 0.21 & 0.4694 & 0.03 & 0.02 & 0.442 \\
\textit{using CSP} & \textbf{0.71} & \textit{\textbf{0.73}} & \textbf{0.4049} & \textbf{0.69} & \textbf{0.72} & \textbf{0.417} & \textbf{0.72} & \textbf{0.74} & \textbf{0.3948} & \textbf{0.78} & \textbf{0.81} & \textbf{0.3951} & \textbf{0.65} & \textbf{0.71} & \textbf{0.4} \\
InternVL2 & 0.78 & 0.74 & 0.0698 & 0.71 & 0.69 & 0.1333 & 0.66 & 0.65 & 0.1846 & 0.41 & 0.21 & 0.2501 & 0.23 & 0.18 & 0.309 \\
\textit{using CSP} & \textbf{0.79} & \textit{\textbf{0.74}} & \textbf{0.0642} & \textbf{0.79} & \textbf{0.73} & \textbf{0.0888} & \textbf{0.78} & \textbf{0.73} & \textbf{0.1285} & \textbf{0.72} & \textbf{0.68} & \textbf{0.2246} & \textbf{0.79} & \textbf{0.82} & \textbf{0.2932} \\
Phi3.5-V & 0.48 & 0.35 & 0.1798 & 0.28 & 0.28 & 0.3768 & 0.28 & 0.28 & 0.3791 & 0.25 & 0.22 & 0.3953 & 0.19 & 0.1 & 0.3424 \\
\textit{using CSP} & \textbf{0.69} & \textbf{0.7} & \textbf{0.095} & \textbf{0.69} & \textbf{0.7} & \textbf{0.2267} & \textbf{0.64} & \textbf{0.67} & \textbf{0.2399} & \textbf{0.54} & \textbf{0.57} & \textbf{0.2878} & \textbf{0.4} & \textbf{0.25} & \textbf{0.3307} \\ \bottomrule
\end{tabular}%
}
\caption{Evaluation of verbalized confidence alignment with correctness across POPE and AMBER datasets. Accuracy (Acc) and F1 Score (F1) are higher-the-better (\(\uparrow\)), while Expected Calibration Error (ECE) is lower-the-better (\(\downarrow\)). These metrics do not measure dataset performance but rather assess the model’s ability to express confidence in alignment with correctness. Our proposed method consistently improves confidence calibration across all models and settings.}
\label{Calibration}
\end{table*}

\begin{figure*}[tbp]
    \centering
    \includegraphics[width=1\linewidth]{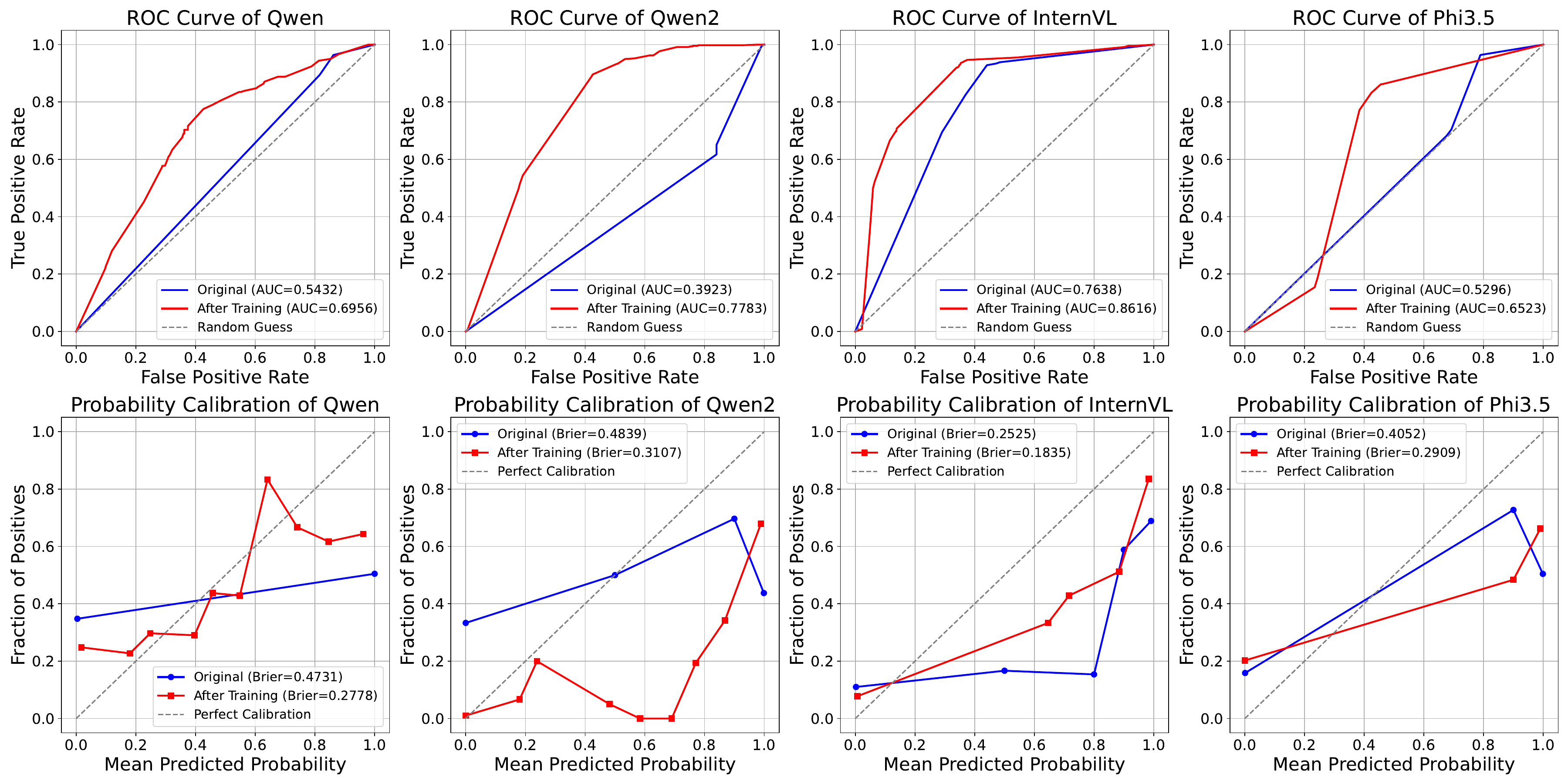}
    \caption{ROC curves (top row) and probability calibration plots (bottom row) on the AMBER attribute dataset, comparing their performance before and after applying our proposed confidence calibration method. The ROC curves illustrate improved true positive rates (higher AUC values) after training, while the probability calibration plots indicate better alignment between predicted confidence and correctness (lower Brier Scores). }
    \label{fig:calibration_amber}
\end{figure*}

Moreover, as the results shown in Figure~\ref{fig:calibration_amber} from AMBER attribute dataset, our approach consistently improves both the Brier Score and the AUC across all tested models, underscoring more accurate confidence estimation and stronger separability between correct and incorrect predictions. 
From the calibration plots of Brier Score, we observe that each model’s calibration curve shifts closer to the diagonal “perfect calibration” line after training, indicating that predicted probabilities better match the actual likelihood of correctness. As a result, the Brier Scores decrease substantially across models, e.g., Qwen-VL decrease from 0.4731 to 0.2778, reflecting reduced mean squared error between predicted probabilities and binary outcomes.
Simultaneously, the ROC Curves show higher AUC, meaning the models after calibration separate true and false positives more effectively for a wide range of confidence thresholds. These joint gains on both Brier Score and AUC confirm that our perturbation-based preference training leads to better verbalized confidence calibration and more reliable confidence judgments, ultimately making the VLMs more trustworthy for multimodal tasks. Additional results for other datasets can be found in the appendix.

\subsubsection{Ablation Experiments}
We conduct an ablation study to evaluate the individual contributions of each component in our object-level verbalized confidence calibration framework. Specifically, we explore four settings: (1) SFT only: using only supervised fine-tuning, (2) SimPO only: applying preference optimization to the base model without SFT, (3) Global Noise: applying Gaussian noise uniformly across the entire image, without focusing on object regions, and (4) Original RLAIF: applying both SFT and SimPO on the original RLAIF dataset without any visual perturbation. We compare these ablations against our full approach, which incorporates semantic mask perturbation, SFT, and SimPO jointly.
\begin{figure}[!htbp]
    \centering
    \includegraphics[width=1\linewidth]{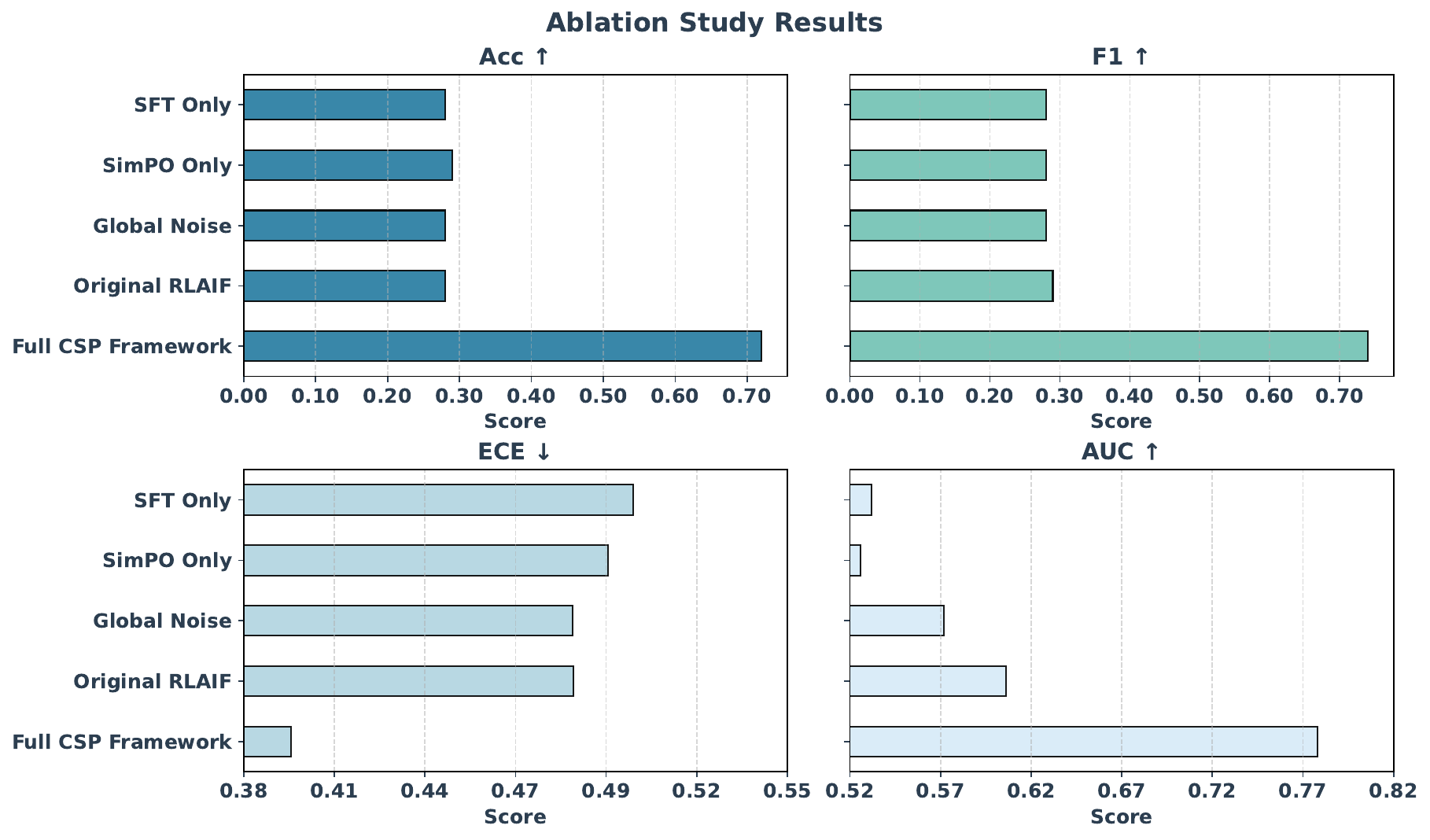}
    \caption{\textbf{Ablation results} for different variants of our method under POPE adversarial of model Qwen2}
    \label{fig:ablation}
\end{figure}
Figure~\ref{fig:ablation} summarizes the performance of each ablation across representative metrics for verbalized confidence calibration. 

\noindent\textbf{SFT Only vs.\ Full Method.} Fine-tuning the model with our perturbation-based dataset alone (without SimPO) yields moderate improvements, confirming the value of learning from controlled object-level uncertainty. However, it remains notably behind the performance of our full CSP framework. This suggests that while learning from the perturbed images is beneficial, the model also needs preference optimization to robustly align its outputs and confidence estimation.

\noindent\textbf{SimPO Only vs.\ Full Method.} Directly applying SimPO to the base model without SFT on perturbed data shows nearly no gains. Without the exposure to semantic perturbations, preference optimization alone struggles to calibrate the model under visual uncertainty.

\noindent \textbf{Global Noise vs.\ Mask-Based Perturbation.} Replacing our object-centric perturbation with uniform global noise fails to improve calibration performance, indicating that indiscriminate perturbation does not meaningfully simulate visual uncertainty. This underscores the importance of masking only the key objects: local, object-centric perturbations more realistically simulate the uncertainty conditions that VLMs encounter, enabling finer control and better confidence calibration.

\noindent \textbf{Original RLAIF vs.\ Perturbation-Based Data.} Finally, using just the original RLAIF dataset for SFT and SimPO does not have improvement. Indeed, the added supervision and varied noise conditions in our semantic perturbation dataset appear crucial for learning robust, multimodal confidence cues.

Altogether, the ablation results emphasize that both object-level mask-based perturbation and preference optimization are essential for achieving robust verbalized confidence calibration. Semantic perturbations expose the model to diverse and realistic uncertainty scenarios grounded in object visibility, while SimPO fine-tunes how the model ranks and verbalizes confidence in alignment with correctness under visual ambiguity.
\begin{figure}[t]
    \centering
    \includegraphics[width=1\linewidth]{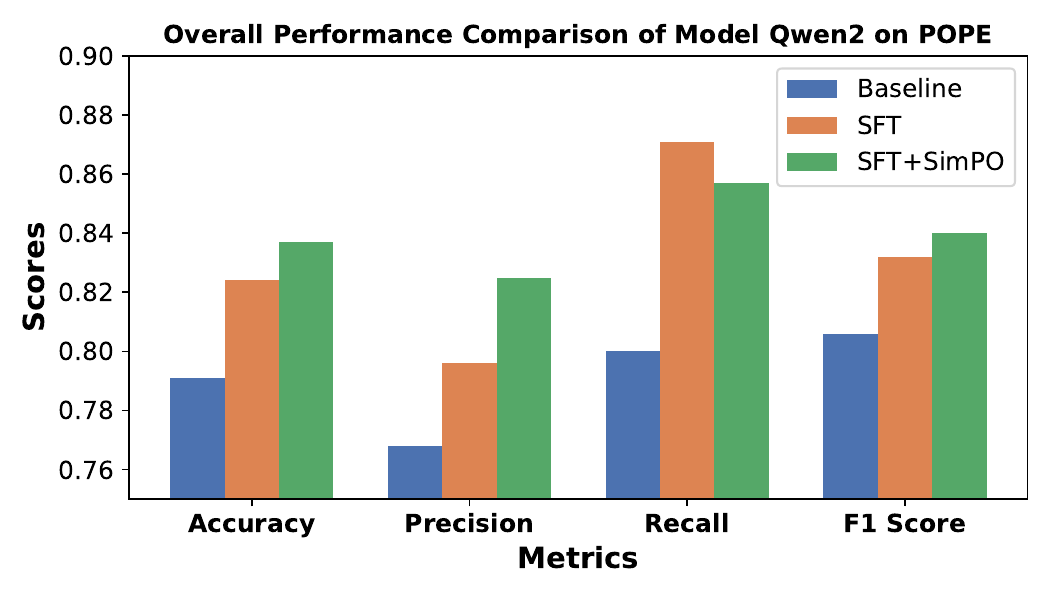}
    \caption{Comparison of Accuracy, Precision, Recall, and F1 Score across different model configurations.}
    \label{fig:performance}
\end{figure}
\begin{table*}[tbp]
\centering
\resizebox{\textwidth}{!}{%
\begin{tabular}{@{}ccccccccccccc@{}}
\toprule
Model & \multicolumn{6}{c}{POPE} & \multicolumn{4}{c}{AMBER} \\ \midrule
 & \multicolumn{2}{c}{Random} & \multicolumn{2}{c}{Popular} & \multicolumn{2}{c}{Adversarial} & \multicolumn{2}{c}{Attribute} & \multicolumn{2}{c}{Relation} \\
 & \textit{Spearman $\rho$} & \textit{Kendall $\tau$} & \textit{Spearman $\rho$} & \textit{Kendall $\tau$} & \textit{Spearman $\rho$} & \textit{Kendall $\tau$} & \textit{Spearman $\rho$} & \textit{Kendall $\tau$} & \textit{Spearman $\rho$} & \textit{Kendall $\tau$} \\ \midrule
Qwen-VL & 0.06 & 0.05 & 0.11 & 0.09 & 0.07 & 0.06 & 0.16 & 0.12 & -0.02 & -0.02 \\
\textit{using CSP} & \textbf{0.16} & \textit{\textbf{0.11}} & \textbf{0.14} & \textbf{0.09} & \textbf{0.14} & \textbf{0.1} & \textbf{0.29} & \textbf{0.2} & \textbf{0.09} & \textbf{0.06} \\
Qwen2-VL & 0.26 & 0.21 & 0.13 & 0.11 & 0.11 & 0.09 & -0.1 & -0.08 & -0.08 & -0.07 \\
\textit{using CSP} & \textbf{0.33} & \textit{\textbf{0.25}} & \textbf{0.29} & \textbf{0.22} & \textbf{0.37} & \textbf{0.27} & \textbf{0.53} & \textbf{0.39} & \textbf{0.22} & \textbf{0.16} \\
InternVL2 & 0.78 & 0.63 & 0.75 & 0.6 & 0.7 & 0.56 & 0.49 & 0.39 & 0.43 & 0.35 \\
\textit{using CSP} & \textbf{0.85} & \textit{\textbf{0.68}} & \textbf{0.83} & \textbf{0.66} & \textbf{0.82} & \textbf{0.65} & \textbf{0.76} & \textbf{0.6} & \textbf{0.61} & \textbf{0.45} \\
Phi3.5-V & 0.55 & 0.45 & 0.39 & 0.31 & 0.36 & 0.29 & 0.17 & 0.13 & 0.24 & 0.19 \\
\textit{using CSP} & \textbf{0.6} & \textbf{0.48} & \textbf{0.57} & \textbf{0.45} & \textbf{0.54} & \textbf{0.42} & \textbf{0.38} & \textbf{0.28} & \textbf{0.38} & \textbf{0.29} \\ \bottomrule
\end{tabular}%
}
\caption{Spearman’s (\(\rho\)) and Kendall’s (\(\tau\)) correlations between internal and verbalized confidence across models and datasets. Higher values indicate better alignment. Our calibration method consistently improves performance.}
\label{probability-confidence alignment}
\end{table*}

\subsubsection{Analysis Experiments}
Our proposed CSP framework not only improves object-level verbalized confidence calibration, but also preserves and in many cases enhances the overall task performance of the VLMs. Below, we analyze two key aspects of our results: (1) the positive impact of benchmark overall performance and (2) the improvement in confidence-probability alignment. 

\noindent \textbf{CSP Enhances Calibration Without Sacrificing Predictive Accuracy}
A common concern in confidence calibration is whether improving verbalized confidence comes at the cost of task performance. Our results show that this is not the case, i.e. our approach does not degrade overall model performance and even enhances key evaluation metrics.
As illustrated in Figure~\ref{fig:performance} of POPE and \ref{tab:mme} of MME\cite{MME}, the bar chart compares key metrics across three configurations: the base vision-language model without additional calibration steps, the model after supervised fine-tuning on the perturbation-augmented dataset, and the model further enhanced through preference optimization.

From the results, we see that both SFT and SFT+SimPO outperform the baseline in all metrics. 
This performance gain is largely due to the semantic perturbation embedded in our dataset. 
By applying perturbations specifically to key semantic regions, the model learns to associate visual uncertainty with corresponding confidence levels while preserving task-relevant features. 
As a result, the model becomes better calibrated in its verbalized confidence and simultaneously maintains or improves its accuracy in object-centric prediction tasks.

\begin{figure}[tbp]
    \centering
    \includegraphics[width=1\linewidth]{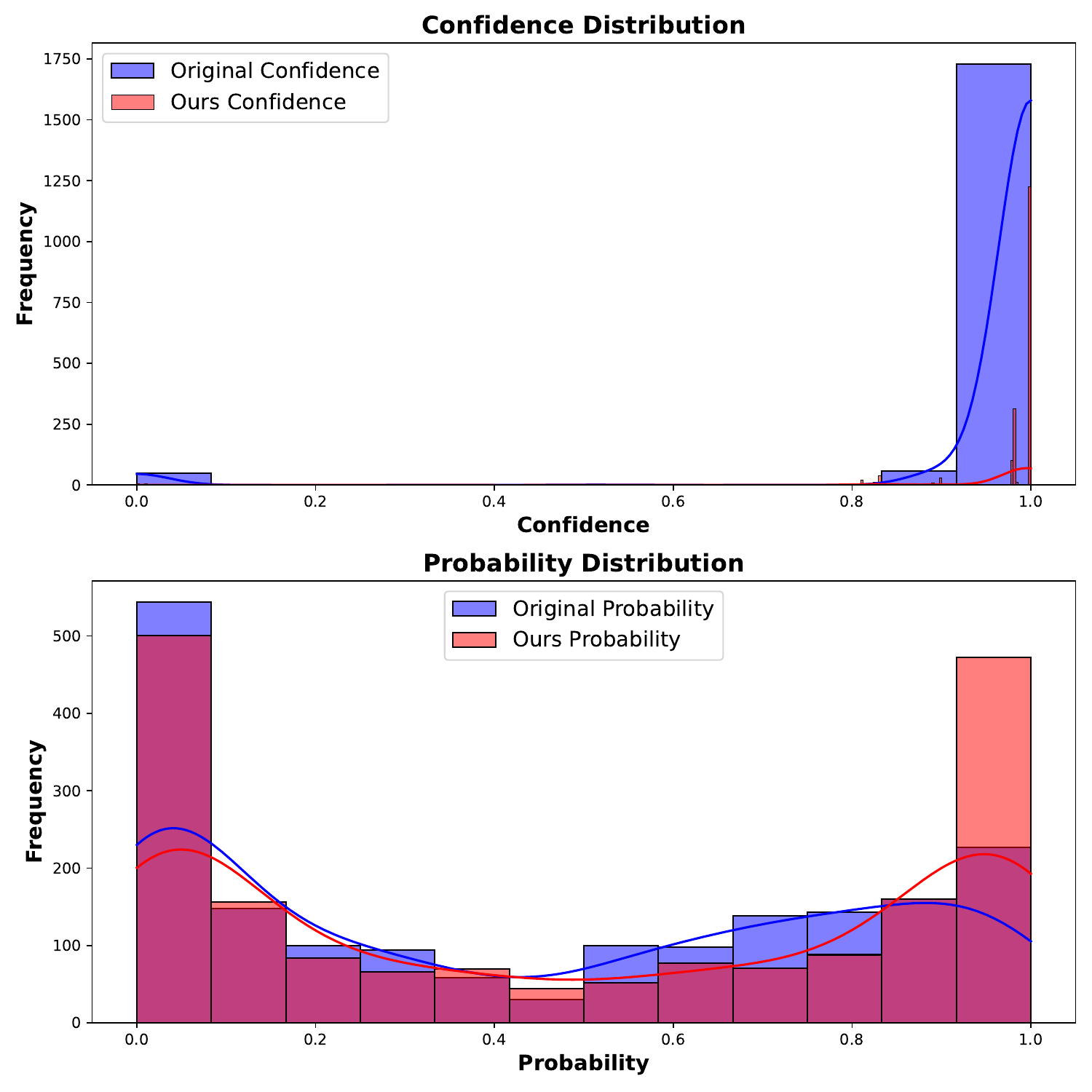}
    \caption{Comparison of token-level probability distributions before (left) and after (right) applying our method.}
    \label{fig:token}
\end{figure}

\noindent \textbf{Indirect Improvement in Internal Confidence-Probability Alignment}
Although our training process does not explicitly constrain token-level internal confidence, Table~\ref{probability-confidence alignment} and Figure~\ref{fig:token} reveal a notable improvement in the alignment between internal probabilities and verbalized confidence. Before calibration, the model exhibited strong overconfidence, assigning excessively high probabilities even to uncertain responses. After fine-tuning with semantic perturbations and preference optimization, the model produces more balanced probability distributions, reducing misleadingly high-confidence predictions for uncertain object-level responses.

We hypothesize that this improvement arises from multiple indirect effects:
(i) semantic perturbations expose the model to controlled object-level uncertainty, prompting more realistic confidence modulation;
(ii) preference optimization reinforces confidence ranking consistency; and
(iii) reduced multimodal bias encourages the model to rely more on visual evidence rather than linguistic priors.
These findings suggest that improved modeling of visual uncertainty not only benefits verbalized confidence, but also indirectly enhances the alignment between internal token-level probabilities and correctness—contributing to overall model trustworthiness.

\section{Conclusion}
This work addresses a key gap in vision-language learning: the lack of reliable object-level verbalized confidence calibration. We introduce CSP, a structured framework that simulates controllable visual uncertainty via semantic perturbations and teaches models to align expressed confidence with visual ambiguity. 
Our findings show that even without directly supervising internal probabilities, CSP improves confidence correctness alignment and enhances overall model reliability. These results highlight the broader potential of object-grounded uncertainty modeling—not only for reducing hallucination, but also for enabling more interpretable and trustworthy multimodal reasoning.
\section*{Limitations}
While our approach demonstrates consistent improvements in object-level verbalized confidence calibration, we acknowledge several areas that merit further exploration:

\begin{itemize}[topsep=0pt,itemsep=2pt,leftmargin=1em] 
\item \textbf{Model Scale.} Our experiments are conducted on mid-sized vision-language models, balancing empirical depth and computational feasibility. While the observed trends are encouraging, assessing whether the benefits of CSP scale with larger models remains an open direction for future work.

\item \textbf{Object-Level Focus.} This work deliberately centers on object-centric perturbations, which provide a controllable and interpretable foundation for modeling localized visual uncertainty. However, broader forms of uncertainty—such as those involving relational reasoning, background context, or temporal ambiguity—may require extending this framework to more diverse semantic levels.

\item \textbf{Task and Domain Generalization.} While CSP shows robust results on hallucination-oriented benchmarks like POPE and AMBER, its applicability to other types of tasks (e.g., reasoning, open-ended generation) and domains (e.g., medical or scientific imaging) remains to be validated. We view this as an opportunity to extend CSP toward a broader range of multimodal scenarios.

\end{itemize}

These limitations reflect the current scope of the study and also suggest rich opportunities for extending semantic perturbation-based calibration to a broader range of models, modalities, and deployment contexts.
\bibliography{reference}
\appendix
\section{Prompt Templates}
In this section, we present the full prompts in various aspects of our experiments.
\subsection{Key Object Region Extraction Prompt}
we use the following prompts when generating the key object region description for the constructed datasets.
\begin{tcolorbox}[colback=gray!15, colframe=black, arc=3mm, boxrule=0.8pt, width=\linewidth]
Extract the single most important keyword (a noun or object) from each of the following question-answer pairs. Provide only one keyword.

Example 1:
Question: What kind of potato chips are on the plate?
Answer: There are some light yellow thin slice-shaped potato chips in this plate, which look very crispy.
Keyword: potato chips

Example 2:
Question: What color is the car parked outside the house?
Answer: The car parked outside is a bright red sedan.
Keyword: car

Example 3:
Question: What kind of fruits are in the basket?
Answer: The basket contains fresh green apples and ripe yellow bananas.
Keyword: fruits

Now, using the following question and answer, extract one most important keyword. Just output the keyword directly.

Question: {\textbf{\textit{Question}}}

Answer: {\textbf{\textit{Answer}}}

Keyword:
\end{tcolorbox}

\subsection{Prompts for Evaluating Verbalized Confidence}
For the following prompt, a few-shot prompt may be used in some models.
\begin{tcolorbox}[colback=gray!15, colframe=black, arc=3mm, boxrule=0.8pt, width=\linewidth]
"A Language model was asked: \{\textbf{\textit{Query}}\}, Model's answer was \{{\textbf{\textit{Response}}}\} Analyze its answer. How certain are you about model’s answer from 1\%-100\% ? "
\end{tcolorbox}
\section{Additional Experiment Details}
\subsection{Implementation Details}
We use SWIFT \cite{swift} framework for the training procedure of our method. Details of hyperparameters can be seen below. For hyperparameters not explicitly mentioned, we use the default settings.
\subsubsection{Hyperparameters used for supervised fine-tuning}
\begin{itemize}
    \item \textbf{SFT type:} Full
    \item \textbf{Batch size:} 2
    \item \textbf{gradient checkpointing:} True
    \item \textbf{gradient accumulation steps:} 8
    \item \textbf{Number of epochs:} 1
\end{itemize}
\subsubsection{Hyperparameters used for preference optimization}
\begin{itemize}
    \item \textbf{RLHF type:} SimPO
    \item \textbf{Batch size:} 1
    \item \textbf{gradient checkpointing:} True
    \item \textbf{gradient accumulation steps:} 16
    \item \textbf{Number of epochs:} 1
    \item \textbf{$\beta$:} 2.0
    \item \textbf{$\gamma_{simpo}$:} 1.0
    \item \textbf{$\alpha_{cpo}$:} 0.0
    \item \textbf{warm-up ratio} 0.03
\end{itemize}
\subsection{Dataset License}
In this section, we list the licenses of the datasets
we used in this paper. We used the datasets for
research purposes as allowed by the corresponding
licenses and consistent with the intended use.

\textbf{POPE} \cite{POPE}: MIT License. We donwloaded the data from \href{https://huggingface.co/datasets/lmms-lab/POPE}{POPE}.

\textbf{AMBER} \cite{AMBER}: Apache License. We donwloaded the data from \href{https://github.com/junyangwang0410/AMBER/blob/master/}{AMBER}.

\subsection{Computation Requirements}
We ran our experiments on a server with $2\times$ AMD EPYC 7513 32-Core Processor and $4\times$ NVIDIA A100-SXM4-80GB and 1T RAM.
\section{Additional Results}
\subsection{Additional Calibration Results}
We illustrate the additional results of Brier Score and ECE of POPE dataset and AMBER relation dataset.
\begin{figure*}[tbp]
    \centering
    \includegraphics[width=1\linewidth]{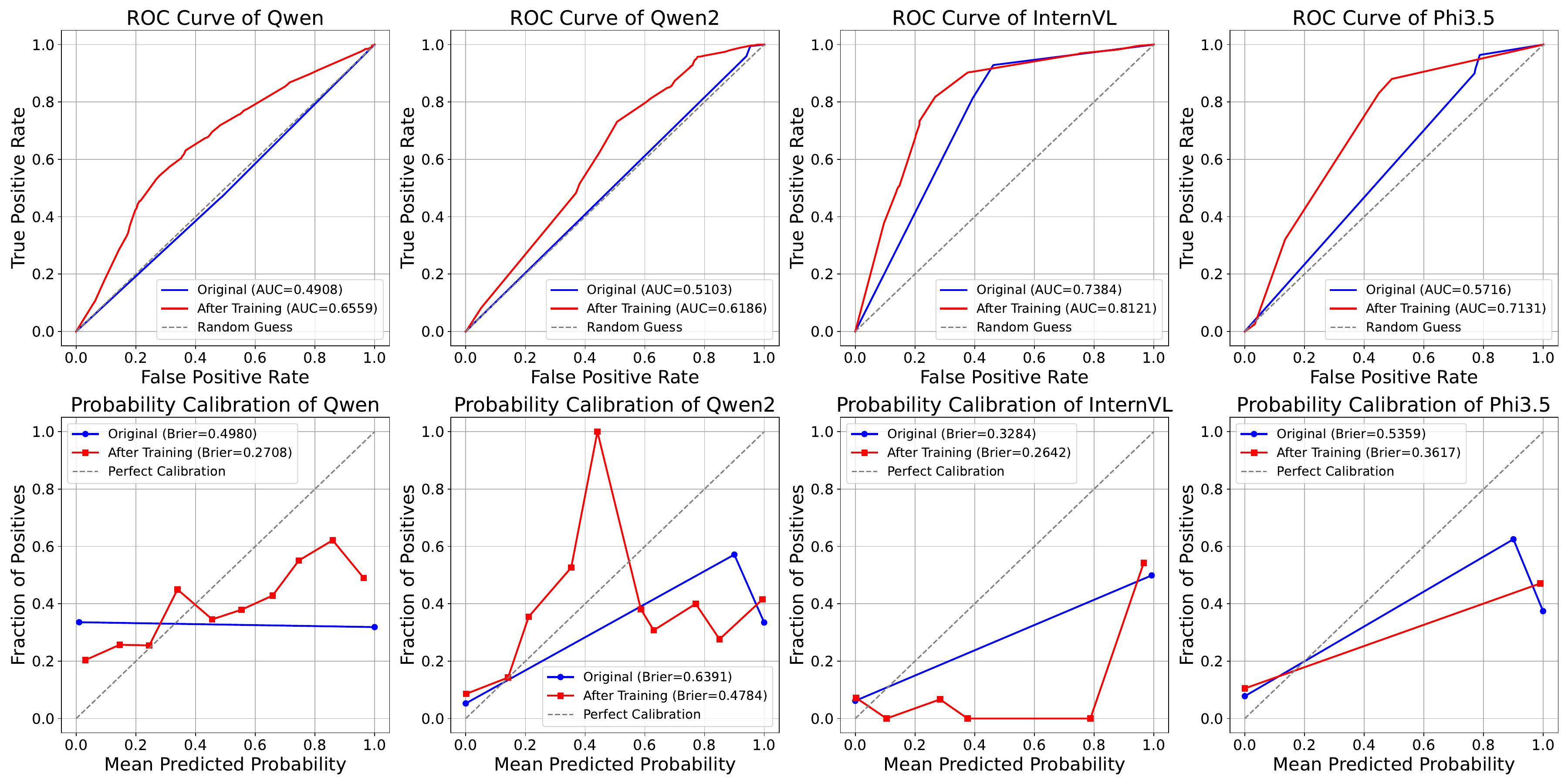}
    \caption{ROC curves (top row) and probability calibration plots (bottom row) on the POPE adversarial dataset.}
\end{figure*}
\begin{figure*}[tbp]
    \centering
    \includegraphics[width=1\linewidth]{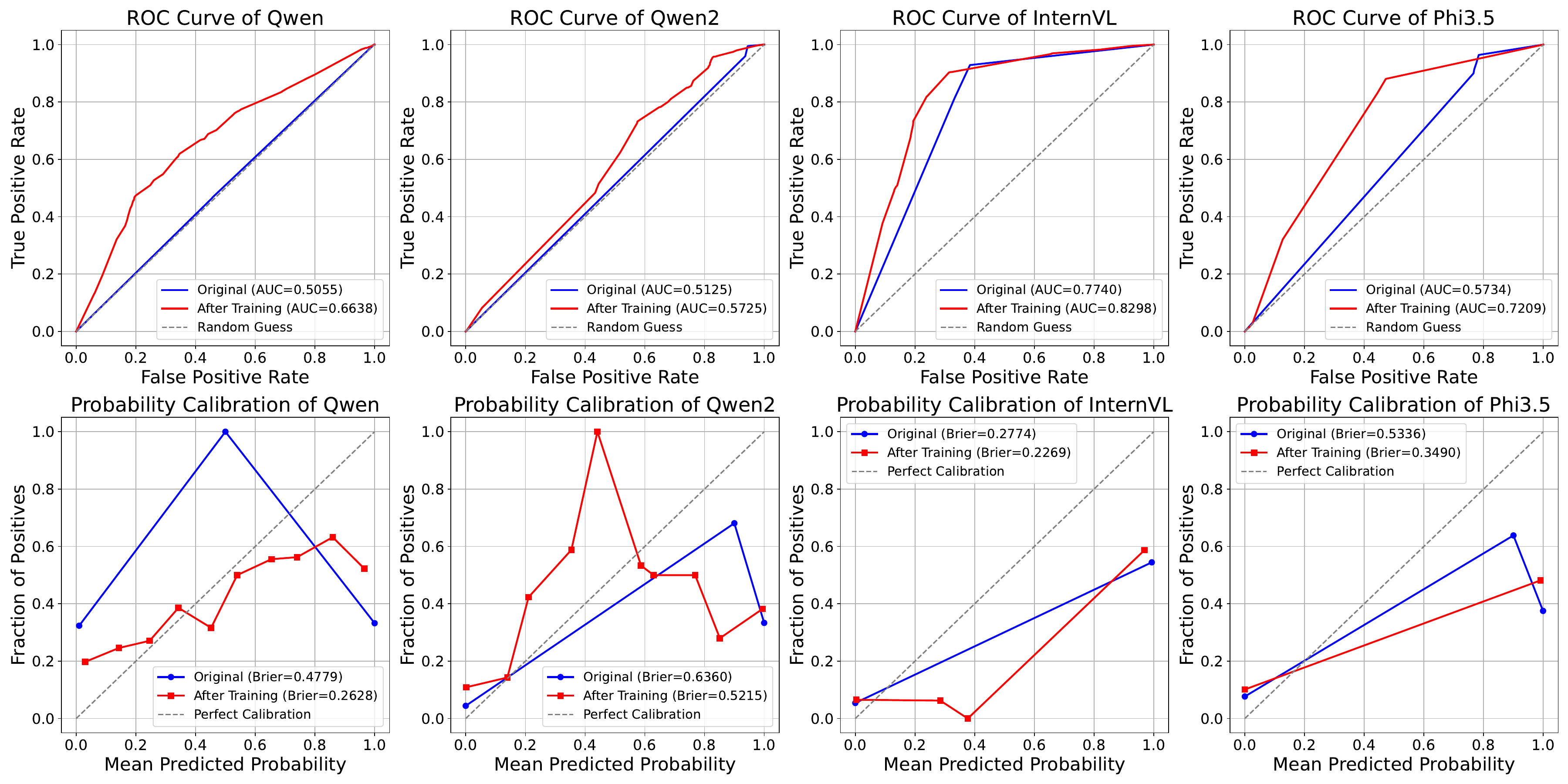}
    \caption{ROC curves (top row) and probability calibration plots (bottom row) on the POPE popular dataset.}
\end{figure*}
\begin{figure*}[tbp]
    \centering
    \includegraphics[width=1\linewidth]{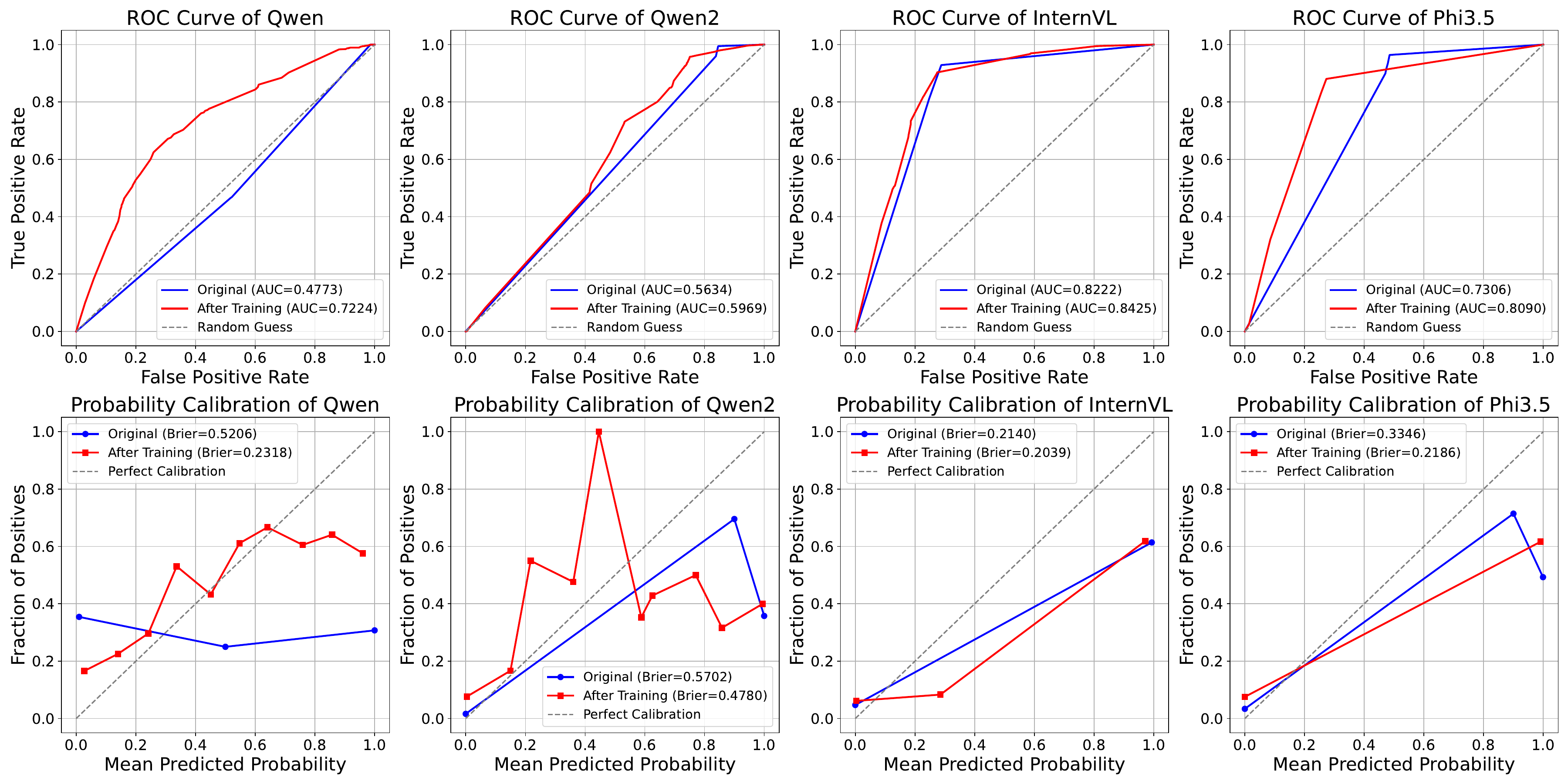}
    \caption{ROC curves (top row) and probability calibration plots (bottom row) on the POPE random dataset.}
\end{figure*}
\begin{figure*}[tbp]
    \centering
    \includegraphics[width=1\linewidth]{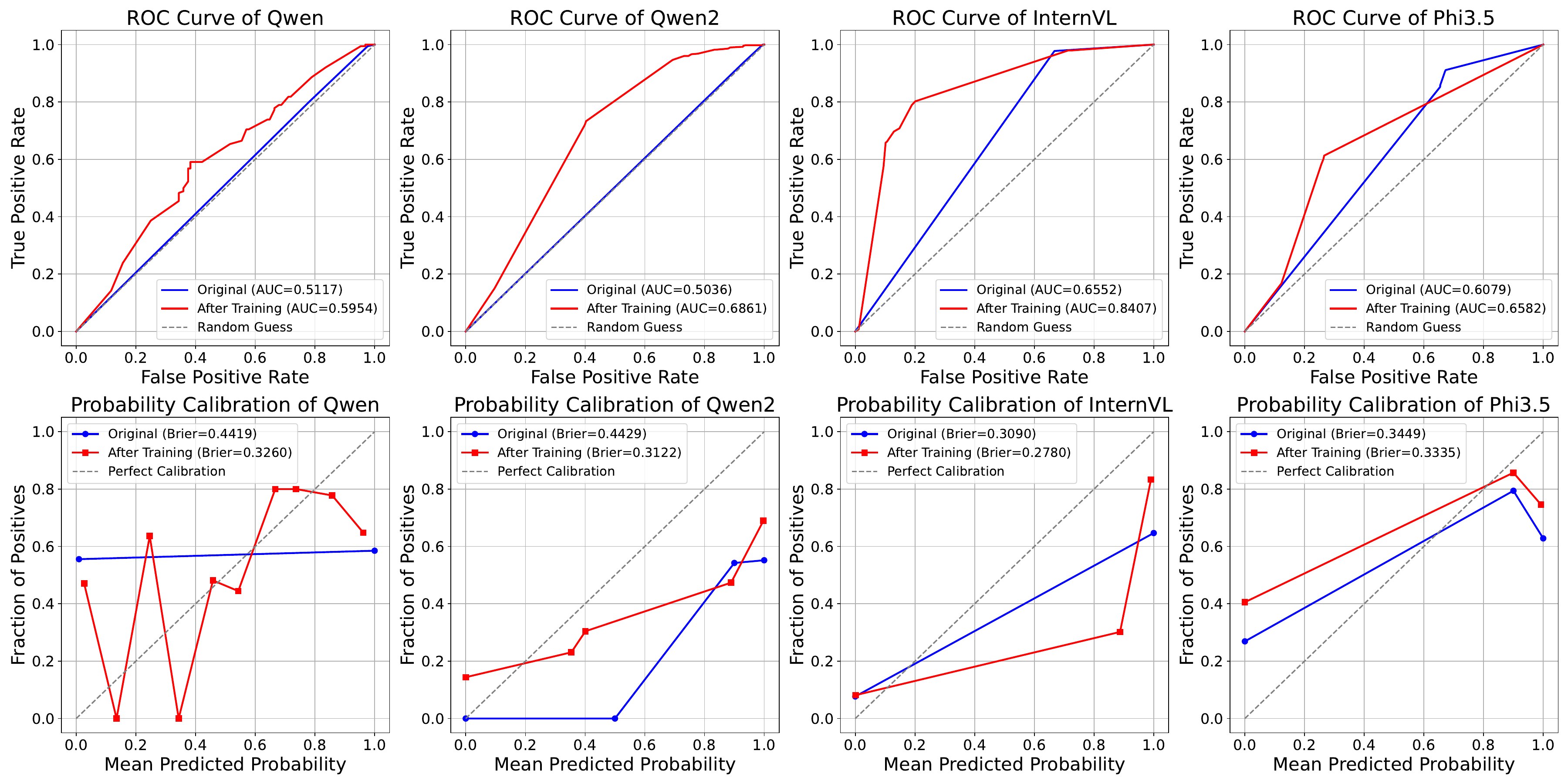}
    \caption{ROC curves (top row) and probability calibration plots (bottom row) on the AMBER relation dataset.}
\end{figure*}
\subsection{Additional Overall Performance on AMBER datasets}
To support our claim regarding the general performance improvements of CSP beyond the POPE dataset, we include additional quantitative results on the AMBER benchmark. Specifically, we evaluate both the \textbf{attribute} and \textbf{relation} subsets of AMBER, which require models to distinguish fine-grained object attributes and spatial/relational concepts under contrastive setups.

We compare three settings: the base vision-language model, the model after supervised fine-tuning (SFT) on the perturbation-augmented dataset, and the model further trained with preference optimization (SFT+SimPO). The results are summarized below.

\begin{table}[t]
\centering
\small
\setlength{\tabcolsep}{3pt}
\begin{tabular}{lcccc}
\toprule
\textbf{Model} & \textbf{Accuracy} & \textbf{Precision} & \textbf{Recall} & \textbf{F1 Score} \\
\midrule
Baseline       & 78.2 & 81.3 & 73.2 & 77.0 \\
SFT            & 80.3 & 82.3 & 77.2 & 79.7 \\
SFT+SimPO      & 80.9 & 77.5 & 87.0 & 82.0 \\
\bottomrule
\end{tabular}
\caption{Performance on AMBER Attribute subset.}
\label{tab:amber-attribute}
\end{table}

\begin{table}[t]
\centering
\small
\setlength{\tabcolsep}{3pt}
\begin{tabular}{lcccc}
\toprule
\textbf{Model} & \textbf{Accuracy} & \textbf{Precision} & \textbf{Recall} & \textbf{F1 Score} \\
\midrule
Baseline       & 70.1 & 64.9 & 68.6 & 66.7 \\
SFT            & 72.0 & 65.7 & 75.3 & 70.2 \\
SFT+SimPO      & 73.6 & 66.2 & 80.8 & 72.8 \\
\bottomrule
\end{tabular}
\caption{Performance on AMBER Relation subset.}
\label{tab:amber-relation}
\end{table}
\subsection{Evaluation on MME Benchmark}
To address concerns regarding potential trade-offs in general multimodal capabilities, we evaluate our CSP framework on the MME benchmark. MME is a comprehensive multimodal evaluation suite covering both perception and reasoning abilities across 14 subtasks, including existence, count, color, landmark recognition, OCR, commonsense reasoning, and more.

We report results for multiple vision-language models (VLMs) before and after applying CSP. The results are shown in Table~\ref{tab:mme}, covering overall accuracy and macro-averaged F1 score.
\begin{table}[t]
\centering
\small
\setlength{\tabcolsep}{4pt}
\begin{tabular}{lcc}
\toprule
\textbf{Model} & \textbf{Accuracy} & \textbf{F1 Score} \\
\midrule
Qwen           & 0.8037 & 0.7250 \\
Qwen + CSP     & 0.8130 & 0.7207 \\
Qwen2          & 0.8602 & 0.8523 \\
Qwen2 + CSP    & 0.8787 & 0.8779 \\
InternVL       & 0.8450 & 0.8450 \\
InternVL + CSP & 0.8610 & 0.8384 \\
Phi3.5         & 0.7839 & 0.7537 \\
Phi3.5 + CSP   & 0.7818 & 0.7402 \\
\bottomrule
\end{tabular}
\caption{Performance on MME benchmark before and after applying CSP.}
\label{tab:mme}
\end{table}
As shown, VLMs maintain or slightly improve their performance on general multimodal tasks after applying CSP. We attribute this to the relatively small amount of fine-tuning data used compared to the models' large-scale pretraining. These results further confirm that our approach improves verbalized confidence calibration without compromising overall multimodal competence.
\end{document}